\documentclass[10pt,twocolumn,letterpaper]{article}

\usepackage{wacv}
\usepackage{times}
\usepackage{epsfig}
\usepackage{graphicx}
\usepackage{amsmath}
\usepackage{amssymb}
\usepackage{booktabs}
\usepackage{multirow, microtype}
\usepackage{authblk}
\usepackage[accsupp]{axessibility}

\def\wacvPaperID{992} 

\wacvalgorithmstrack   

\wacvfinalcopy 

\begin{document}

\title{DyAnNet: A Scene Dynamicity Guided Self-Trained Video Anomaly Detection Network}

\author[1]{\small Kamalakar Vijay Thakare}
\author[1]{\small Yash Raghuwanshi}
\author[1]{\small Debi Prosad Dogra}
\author[2,3]{\small Heeseung Choi}
\author[2,3]{\small Ig-Jae Kim}

\affil[1]{\footnotesize Indian Institute of Technology, Bhubaneswar, Odisha, 752050, India}
\affil[2]{\footnotesize Artificial Intelligence and Robotics Institute, Korea Institute of Science and Technology, Seoul 02792, Republic of Korea}
\affil[3]{\footnotesize Yonsei-KIST Convergence Research Institute, Yonsei University, Seoul 03722, Republic of Korea}
\affil[ ]{\textit {\{tkv15, yr15, dpdogra\}@iitbbs.ac.in, \{hschoi, drjay\}@kist.re.kr}}


\maketitle
\thispagestyle{empty}

\begin{abstract}
Unsupervised approaches for video anomaly detection may not perform as good as supervised approaches. However, learning unknown types of anomalies using an unsupervised approach is more practical than a supervised approach as annotation is an extra burden. In this paper, we use isolation tree-based unsupervised clustering to partition the deep feature space of the video segments. The RGB- stream generates a pseudo anomaly score and the flow stream generates a pseudo dynamicity score of a video segment. These scores are then fused using a majority voting scheme to generate preliminary bags of positive and negative segments. However, these bags may not be accurate as the scores are generated only using the current segment which does not represent the global behavior of a typical anomalous event. We then use a refinement strategy based on a cross-branch feed-forward network designed using a popular I3D network to refine both scores. The bags are then refined through a segment re-mapping strategy. The intuition of adding the dynamicity score of a segment with the anomaly score is to enhance the quality of the evidence. The method has been evaluated on three popular video anomaly datasets, i.e., UCF-Crime, CCTV-Fights, and UBI-Fights. Experimental results reveal that the proposed framework achieves competitive accuracy as compared to the state-of-the-art video anomaly detection methods.  
\end{abstract}

\section{Introduction}
\label{section:introduction}
Video Anomaly Detection (VAD) imposes a critical requirement in visual surveillance. Generally, video anomaly detection task covers a large spectrum including road traffic monitoring~\cite{street_scene,Shah2018CADP:Analysis}, violence detection~\cite{150fps,angry_crowd,cctv_fights}, human behaviour~\cite{driver_dataset,pose_clustering_anomaly, human_skeleton_anomaly}, crowd monitoring~\cite{pidlnet,crowd_anomaly}, etc. Visual surveillance is primarily done by public and private agencies on a large scale. Hence researchers easily get humongous data analytic task while analyzing and annotating large video data. Moreover, recent existing video anomaly detection methods~\cite{mist,pose_clustering_anomaly,human_skeleton_anomaly,street_scene,sultani_chen_shah_2018,jiaxingzhongnannanliweijiekong2019,Zhu2020Motion-awareDetection} heavily depend on full or weak supervision. However, generating annotations for such huge datasets is labor-intensive and time-consuming.

\begin{figure}
\label{figure:one}
    \centering
    \includegraphics[scale=0.75]{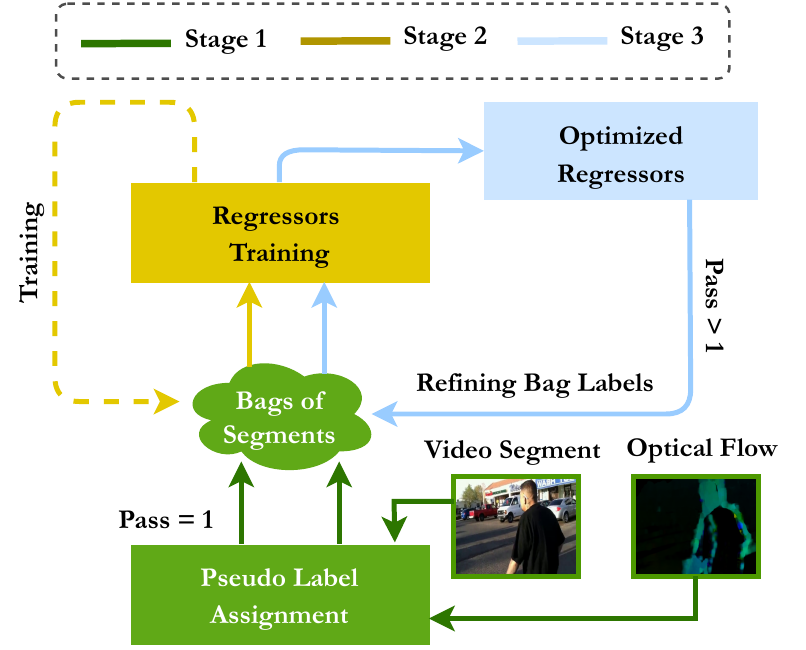}
    \caption{\textbf{Overview}. In the first stage, we obtain low-confidence pseudo labels. In the second stage, we incorporate \textit{iterative learning} to train regressor networks using these labels. After successful training, we replace older labels with more confident labels in the third stage and retrain the regressors. After a few passes, an optimized version of regressors is used to predict the anomaly score.}
\end{figure}

In recent years, unsupervised approaches for video anomaly detection are being outnumbered by supervised or semi-supervised methods. Ravanbaksh~\etal~\cite{ravanbaksh} have trained Generative Adversarial Nets (GANs) for video anomaly detection. Nguyen~\etal~\cite{nguyen} have concatenated appearance and motion encoders and decoders for accomplishing the job. Gong~\etal~\cite{ae_memory} have proposed Memory-augmented Autoencoders (MemAEs) to detect video anomalies. The main advantage of using GANs or AEs is their capability to capture high-level video features. Recently, Doshi~\etal~\cite{continual} have proposed a continual learning framework in which the model incrementally trains as the data arrives without forgetting the learnt (past) information. This type of framework can be feasible in visual surveillance as video data keep coming into the monitoring systems. However, all these approaches suffer a few limitations as follows: (1) in continual learning, a separate mechanism needs to be designed to avoid catastrophic forgetting~\cite{continual}, (2) GANs and AEs are highly vulnerable to unstable training, i.e., a subtle change in data imposes large changes in the labels, thus affecting the normal distribution, (3) most of the state-of-art VAD methods heavily depend on labeled normal/abnormal data, and (4) VAD approaches either utilize appearance-based features or deep features.

To address these limitations, we adopt an iterative learning~\cite{iterative_noisy} mechanism in which models are repeatedly tuned with more refined data during each pass. Moreover, we aim to combine the technical advantages of continual and AEs learning. Our proposed framework combines the power of DNNs with well-justified handcrafted motion features. These spatio-temporal features equipped with low-level motion features help to detect wide range of anomalies. The framework can also be retrained in an end-to-end fashion as input data arrives. The overview of the proposed framework is depicted in Fig.~\ref{figure:one}. It is divided into three stages: i) pseudo label assignment, ii) regressors training, and iii) refinement of labels using optimized regressors. For enabling the regressors to understand subtle anomalies, we have obtained motion features, namely dynamicity score using optical flow. In the first stage, we do not know the actual labels; hence we have obtained intermediate low confidence anomaly labels using OneClassSVM and iForest~\cite{isolation}. We also obtain the dynamicity labels using dynamicity scores. We have trained two regressor networks in the second stage by using the labels generated in the first stage. This is an iterative process to improve the confidence scores. In this way, both regressors are trained over refined labels and they learn discriminating features. The iterative learning approach also ensures that both the regressors learn new distinguish patterns without losing the past information. We have experimentally found that for first few iterations, both regressors gradually learn internal patterns and stabilizes after some iterations. Both regressors are trained independently in parallel. Precisely, in iterative learning, the model is retrained using refined data in each iteration. In this way, the proposed approach do not need any level of supervision. However, some form of supervision is mandatory for continual learning~\cite{continual} or weakly-supervised methods~\cite{nguyen,sultani_chen_shah_2018,jiaxingzhongnannanliweijiekong2019}. These methods consider a video anomalous even if a small segment contains anomaly. In contrast, we identify anomalous segments using dynamicity and anomaly scores estimated using  unsupervised ways, thus eliminating the requirement of supervision. To achieve this, we have made the following contributions:

\begin{itemize}
    \item design an unsupervised end-to-end video anomaly detection framework that uses iterative learning to tune the model using refined labels in each iteration;
    
    \item propose a novel technique to assign intermediate labels in unsupervised scenarios by combining deep features with well-justified motion features and;
    
    \item conduct extensive experiments to understand the effectiveness of the proposed framework with respect to other state-of-the-art methods.
\end{itemize}
The rest of the paper is organized  as follows. In the next section, we present the related work. In Sec.~\ref{section:proposed_method}, we present the proposed framework. Experiments and results are presented in Sec.~\ref{section:experiments}. The conclusions and future works are presented in Sec.~\ref{section:conclusion}.
\section{Related Work}
\label{section:related_work}
Existing work in the Video Anomaly Detection (VAD) domain largely draw motivation from activity recognition and scene understanding~\cite{sultani_chen_shah_2018}. These methods utilize various types of video features, training procedures or both. In this section, we briefly discuss the main categories that are extensively followed in very recent VAD approaches. 

\subsection{Reconstruction-based Approaches}
Several VAD approaches~\cite{autoregression_ae,ae_memory,future_frame_vae,nguyen,fastano,mnad_ae,szyma_wacv22,old_is_gold} employ Autoencoders (AEs), Generative Adversarial Nets (GANs) and their variants under the assumption that the models that are explicitly trained on normal data may not be successful to reconstruct abnormal event as such samples are usually absent in the training set. Park~\etal~\cite{fastano} have used AE to generate cuboids within normal frames using spatial and temporal transformation. Zaheer~\etal~\cite{old_is_gold} have generated good quality reconstructions using the current generator and used the previous state generator to obtain bad quality examples. This way, the new discriminator learns to detect even small distortions in abnormal input. Gong~\etal~\cite{ae_memory} have introduced a memory module to AE and constructed MemAE. This is an improved version of existing AE. Szymanowicz~\etal~\cite{szyma_wacv22} have trained an AE to obtain saliency maps using five consecutive frames and per-pixel prediction error. Ravanbakhsh~\etal~\cite{ravanbaksh} have imposed classic adversarial training using GANs to detect anomalous activity. However, the effectiveness of these approaches is highly dependent on the reconstruction capabilities of the model. Failing which, it may significantly degrade the model's performance.

\begin{figure*}
    \centering
    \includegraphics[scale=0.70]{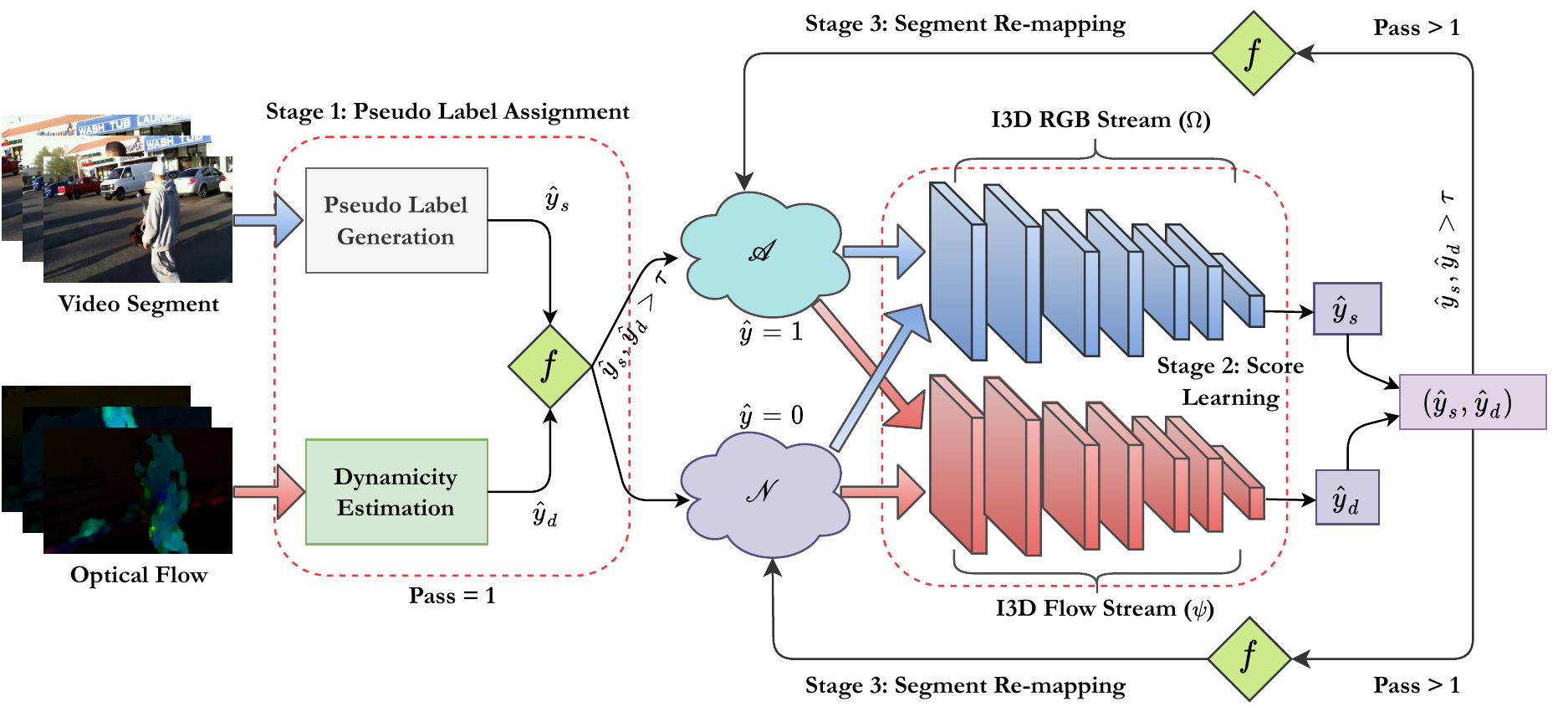}
    \caption{\textbf{DyAnNet}. Architecture of the proposed framework. The whole framework is divided into three stages: (1) Pseudo label assignment, (2) Score learning, and (3) segment re-mapping using refined labels. We have employed iterative learning mechanism to train the regressors $\Omega$ and $\psi$,  and redefined the input bag $\mathcal{A}$ and $\mathcal{N}$ at the end of each pass. We construct a set of optimized regressors obtained through each pass and used it to predict anomaly and dynamicity score of each segment. (Thinner arrows represent the label passing and the thicker arrows are video features, i.e., the blue arrows are raw RGB frames, whereas the red arrows are representing optical flow of the segment.)}
    \label{figure:block_diagram}
\end{figure*}

\subsection{Features-based Approaches}
Primarily, features-based VAD approaches can be categorized by anomaly detection using either handcrafted or deep features. Early attempts have used handcrafted features such as object trajectories~\cite{cvrr,trajectories}, gradients of histograms (HOGs)~\cite{hog}, Hidden Markov Model (HMM)~\cite{hmm}, and appearance-based features~\cite{mahmudul}. However, very recent deep learning approaches~\cite{mist,sultani_chen_shah_2018,jiaxingzhongnannanliweijiekong2019,Zhu2020Motion-awareDetection} have achieved robust results for video anomaly detection. Feng~\etal~\cite{mist} have introduced Self-guided attention during the feature encoding process, Zhu~\etal~\cite{Zhu2020Motion-awareDetection} have injected motion-aware features that increases the recognition capabilities of the classifier. Sultani~\etal~\cite{sultani_chen_shah_2018} addresses anomaly detection problem using weak supervision and following this, ~\cite{jiaxingzhongnannanliweijiekong2019} has used Graph Convolutions Network (GCN). In addition to this, different training mechanisms have been employed such as continual learning~\cite{continual}, adversarial training~\cite{ravanbaksh}, Self-trained~\cite{mist,ordinal}, and active learning~\cite{active} to obtained robust video anomaly detection results.

Even though the aforementioned techniques have achieved decent performance, they still suffer from a few avoidable limitations: (1) they heavily depend on manually labelled normal/abnormal data. However, generating annotations for huge data is time consuming and error prone, (2) due to the absence of universal definition the anomaly events, a few anomalous events that are normal in one context may be regarded abnormal in another context, e.g. marathon run vs. criminal run. These scenarios often lead to unstable training of the AEs and GANs. We have addressed these limitations using iterative learning combined with low and high-level features.

\section{Proposed Method}
\label{section:proposed_method}
We first provides a detailed description of the proposed video anomaly detection framework. Our framework encompasses with the following three stages: (1) pseudo label assignment, (2) anomaly score learning, and (3) segment re-mapping.

\subsection{Overall Architecture}
A high-level architecture of the proposed framework is depicted in Fig.~\ref{figure:block_diagram}. The problem formulation is follows: Assume an input video $(V)$ is divided into a $n$ number of segments such that $V = \{S_{1}, S_{2}, \ldots, S_{n}\}$. The goal is to design a function as given in Eq.~\ref{eq:added1} that generates an anomaly score $y_{s}$ and a dynamicity score $y_{d}$ to predict the label $y \in \{0,1\}$ for each video segment.
\begin{equation}
\label{eq:added1}
    \Theta : V \rightarrow y \in \{0,1\}
\end{equation}

 A positive segment contains anomalous activity and ideally has a higher anomaly and dynamicity score than the normal segments such that $\Theta(S_{i}) > \Theta(S_{j})$, where $S_{i}$ is an anomalous and $S_{j}$ is a normal segment. Note that, no labeled data is available during this training. To tackle this scenario, we have employed iterative learning~\cite{iterative_noisy} and bag formation~\cite{sultani_chen_shah_2018}. First, we have assigned pseudo anomaly scores $\hat{y}_{s}$ and pseudo dynamicity scores $\hat{y}_{d}$ to the video segments $S_i$. These intermediate labels help to form two separate bags $\mathcal{A}$ and $\mathcal{N}$. Here, $\mathcal{A} \subset V$ is bag of positive segments, where $S \in \mathcal{A}$ if $y = 1$ for $S$, which generally has higher $\hat{y}_{s}$ and $\hat{y}_{d}$ values. Similarly, $\mathcal{N} \subset V$ is the bag of negative segments, where $S \in \mathcal{N}$ if $y = 0$ for normal segment $S$ and we expect a lower value for both $\hat{y}_{s}$ and $\hat{y}_{d}$. Note that, $\mathcal{A} \cap \mathcal{N} = \phi$. In the second stage, two separate regressors, e.g. $\Omega$ and $\psi$ have been trained using these pseudo labels. In the third stage, we have used these trained regressors to refine the contents of the bags. A training pass redefines the membership of each segment of a bag. In the next pass, $\Omega$ and $\psi$ are tuned using $\mathcal{A}$ and $\mathcal{N}$. In the subsequent sections, we provide detailed descriptions of the stages.

\subsection{Pseudo Label Assignment}
The training procedure begins with unlabeled data. Hence we don't know $\mathcal{A}$ and $\mathcal{N}$ in the first place. To handle this problem, we initialize $\mathcal{A}$ and $\mathcal{N}$ via generating pseudo anomaly score $\hat{y}_{s}$ and pseudo dynamicity score $\hat{y}_{d}$. To obtain $\hat{y}_{s}$, we have employed OneClassSVM and iForest~\cite{isolation} in combination. Note that, both algorithms run on feature vectors of the video segment. We have extracted segment features using I3D pretrained on Kinetic dataset~\cite{j.carreiraandrewzisserman2017}. OneClassSVM is similar to the SVM algorithm. But, it uses hypersphere to cover all data instances. This algorithm tries to construct the smallest possible hypersphere using the support vectors. All the data instances that lie outside the hypersphere are likely to be anomalies. Let $F = f(S)$ be the feature extraction function of segment $S$. The anomaly score can be defined using Eq.~\ref{equation:one},

\begin{equation}
\label{equation:one}
    d(F) = \max_{F \in V} \delta(c, F)
\end{equation}
where $F$ is the feature point, $c$ is the center of the smallest hypersphere constructed by the SVM, and $\delta$ is the distance function. iForest isolates data instances by randomly selecting any feature and a split value. A tree structure can depict this recursive partitioning; hence the number of partitions is equal to the path length of the data instance up to the root node. The inverse of the path from the root to leaf is the anomaly score of $F$. It is estimated using Eq.~\ref{equation:two},

\begin{equation}
\label{equation:two}
d(F) = 2^{[\frac{- E(l(F))}{g(|F|)}]}
\end{equation}
where $l(F)$ is the path length of $F$, $E(.)$ denotes the average path length of $F$ on $n$ isolation trees, and $g(.)$ is the expected path length for a given sub-sample. We normalize the anomaly scores of each feature point within [0,1] interval and take the average score over $n$ isolation trees to obtain the pseudo anomaly score $\hat{y_{s}}$ of a video segment.

In addition to the anomaly score, we have also obtained the dynamicity score of each segment. The dynamicity of the segment refers to the rate of change in displacement of the pixel over time which is obtained using motion information. It is expected that for a rapidly changing video scene, the dynamicity score is expected to be higher. Let $P_{k}$ represents the coordinate of the $k^{\text{th}}$ pixel of the immediate preceding frame and $M_{k}$ be the estimated position of the pixel obtained using optical flow in the next frame. The displacement of the pixel ($S_{k}$) can be calculated using Eq.~\ref{equation:three},
\begin{equation}
\label{equation:three}
    S_{k} = SAD(P_{k}, M_{k}) 
\end{equation}
where SAD is the sum of absolute difference. We have used absolute displacement to consider movement in any direction to estimate the dynamicity score. Now, the frame-level dynamicity score $D_{i}$ of the  $i^{\text{th}}$ frame is estimated using Eq.~\ref{equation:five},
\begin{equation}
\label{equation:five}
    D_{i} = \frac{1}{m \times n} \sum_{k = 1}^{m \times n} S_{k}
\end{equation}
where $m$ and $n$ represent height and width of the frame. We then obtain the dynamicity scores of all the frames within a segment. It is represented by $[D_{i}, D_{i+1}, \ldots, D_{p-1}]$ assuming there are $p$ number of frames in a segment. We average all frame-level dynamicity scores to obtain a pseudo dynamicity score $\hat{y}_{d}$ of the segment. The score is then normalized within [0,1]. We now assign an intermediate label $\hat{y}$ to a segment using the heuristic presented in Eq. \ref{equation:six}.

\begin{equation}
\label{equation:six} 
    \hat{y} = f(\hat{y}_{s}, \hat{y}_{d}) =
\begin{cases}
  1 & \text{if $\hat{y}_{s}, \hat{y}_{d} > \tau$} \\
  0 & \text{otherwise}
\end{cases}
\end{equation}
Eq.~\ref{equation:six} ensures that the segment with a higher anomaly and dynamicity scores than a predefined threshold $\tau$ should be placed in $\mathcal{A}$ with the intermediate label $\hat{y} = 1$.

\subsection{Learning of Anomaly and Dynamicity Scores}
Ideally, when a score learner feeds with an anomalous segment, it should generate a high anomaly score as compared to a normal segment. However, in the present scenario, the labels are inaccurate due to the absence of ground truths. Moreover, the label of each segment has been decided using anomaly and dynamicity scores. Hence we have carefully designed a function $\Theta$ as given in Eq.~\ref{equation:seven} using two different score learner functions, namely $\Omega$ and $\psi$, 

\begin{equation}
\label{equation:seven}
    \Theta(S) = f(\Omega(Z_{R}), \psi(Z_{F})),
\end{equation}
where $Z_{R}$ represents the RGB frame, $Z_{F}$ denotes the optical flow of the segment $S$,  $\hat{y}_{s} = (\Omega(Z_{R}))$, $\hat{y}_{d} = (\psi(Z_{F}))$, either $S \in \mathcal{A}$ or $S \in \mathcal{N}$ and $f(.)$ is the label mapping function defined in Eq.~\ref{equation:six}. Typical 3D CNNs can be incorporated here to implement $\Omega$ and $\psi$. We have employed RGB and flow modalities of I3D~\cite{j.carreiraandrewzisserman2017} network followed by a 3-layer FCN to implement score learners $\Omega$ and $\psi$, respectively. We train the anomaly score learner $\Omega(Z_{R},\mathbf{W_{\Omega}})$ and dynamicity score learner $\psi(Z_{R}, \mathbf{W_{\psi}})$ using Mean-squared error (MSE) loss, where $\mathbf{W}_{\Omega}$ and $\mathbf{W}_{\psi}$ are trainable weights of $\Omega$ and $\psi$, respectively.

\subsection{Segment Re-mapping via Iterative Learning}
The training procedure begins with the pseudo labels assigned to the segments in the first stage. However, labels are not as correct as the ground truth. In this stage, we aim to fine-tune $\Omega$ and $\psi$ with more accurate labels to achieve stable performance. To achieve this, we have incorporated an iterative learning mechanism. Let $P_{i}$ be the $i^{th}$ pass in which $\mathcal{A}$ and $\mathcal{N}$ have been initialized based on pseudo anomaly and dynamicity scores. We then train $\Omega$ and $\psi$ via MSE loss using these pseudo labels. We have obtained sub-optimized version $\Omega_{P_{i}}$ and $\psi_{P_{i}}$ of both regressors. Lastly, we re-estimate both scores using Eq.~\ref{equation:eight} via these optimized versions of the regressors,

\begin{equation}
\label{equation:eight}
    \hat{y}_{s}^{P_{i+1}} = \Omega_{P_{i}}(Z_{R}) \hspace{2mm} \text{and} \hspace{2mm} \hat{y}_{d}^{P_{i+1}} = \psi_{P_{i}}(Z_{F})
\end{equation}
where $\hat{y}_{s}^{P_{i+1}}$ and $\hat{y}_{d}^{P_{i+1}}$ are new scores obtained through sub-optimized regressors. Now, we use these new scores to refine $\mathcal{A}$ and $\mathcal{N}$ using Eq.~\ref{equation:six} and retrained $\Omega_{P_{i}}$ and $\psi_{P_{i}}$ in the next pass $P_{i+1}$ using a new input batch. In particular, for each pass in iterative learning, we utilize a completely new set of $\mathcal{A}$ and $\mathcal{N}$ and retrain the regressors. We only utilize new scores instead of combining them with older scores because such a mixing without any supervision usually generates erroneous scores. We have empirically found that the proposed approach performs better on popular video anomaly datasets. Finally, each pass generates an optimized version of $\Omega$ and $\psi$ and hence the proposed iterative learning approach results in a set of optimized regressor models.

\subsection{Training and Inference}
We have employed iterative learning to achieve stable performance of the regressors. During the first pass, we have obtained pseudo anomaly and dynamicity scores to initialize $\mathcal{A}$ and $\mathcal{N}$. However, the actual training takes place in the second stage, where two regressor models $\Omega$ and $\psi$ are trained using the pseudo labels. Note that, $\Omega$ and $\psi$ are I3D~\cite{j.carreiraandrewzisserman2017} networks followed by a 3-layer FCN with a single neuron at the end to produce respective scores. Hence we have incorporated MSE loss for network the training as the formulation is recognized as regression rather than a binary classification. In each pass, both regressor networks have been trained using a fixed number of training iterations depending on the number of samples available in the training set. Finally, the sub-optimized version of $\Omega$ and $\psi$ are used to rearrange the content of $\mathcal{A}$ and $\mathcal{N}$ for the next pass.

Each pass in iterative learning outputs an optimized version of $\mathcal{A}$ and $\mathcal{N}$. In the inference stage, we use a set of sub-optimized models to generate optimized anomaly and dynamicity scores. The final score generation can be summarized using Eq.~\ref{equation:nine},

\begin{equation}
\label{equation:nine}
    y_{s} = \sum_{i = 1}^{k} \Omega_{i}(Z_{R}) \hspace{2mm} \text{and} \hspace{2mm} y_{d} = \sum_{i = 1}^{k} \psi_{i}(Z_{F})
\end{equation}
where $k$ is the number of passes. $\Omega_{i}$ and $\psi_{i}$ represent the optimized models obtained after the $i^{\text{th}}$ pass. $y_{s}$ and $y_{d}$ are anomaly and dynamicity scores obtained using $\Omega_{i}$ and $\psi_{i}$, respectively. The output neuron from both the regressors use softmax, hence anomaly and dynamicity scores always fall between [0,1] for an input video segment.
\section{Experiments}
\label{section:experiments}
In this section, we present implementation details, datasets, evaluation metrics, comparisons of the proposed method with recent state-of-the-art VAD methods, qualitative results, ablation experiments, and the effect of training and testing iterations on performance.

\subsection{Implementation Details}
Following~\cite{sultani_chen_shah_2018,Zhu2020Motion-awareDetection,jiaxingzhongnannanliweijiekong2019}, we divide each video into 32 non-overlapping temporal segments. We then extract features from \textit{mixed-5C} layer of the I3D~\cite{j.carreiraandrewzisserman2017} network resulting in 1024D feature components, and feed them to PCA to reduce dimensionality to 100 components. These components have been used to train the OneClassSVM and iForest~\cite{isolation} classifiers to generate pseudo anomaly scores. We have used default parameters of OneClassSVM and iForest~\cite{isolation} given in the scikit-learn during experiments.

We have used SelFlow~\cite{selflow} and Farneback algorithm for optical flow estimation to calculate the dynamicity score of the segment. We have implemented the regressors $\Omega$ and $\psi$ using I3D~\cite{j.carreiraandrewzisserman2017} as a backbone network pre-trained on the Kinetic dataset as recommended in I3D original work. We have replaced FCN layers of I3D with a 3-layer FCN. The first layer contains 512 units, followed by 32 units, and 1 unit at the end to generate the scores. We have also experimented with deeper networks. However, we have not observed significant performance deviation. We have trained the regressors with initial learning rate of 0.005 and AdaGrad optimizer. Following  \cite{Hinami2017JointKnowledge,sultani_chen_shah_2018, jiaxingzhongnannanliweijiekong2019,Zhu2020Motion-awareDetection}, we set $\tau = 0.50$ for comparisons. We have experimented with even lower values of $(\tau)$. Such analysis can be found in supplementary document. The experiments reveal that both regressors get substantially improved only in the first few passes while achieving a stable performance. We have discussed results by varying the number of training iterations and passes in the subsequent sections.

\subsection{Datasets}
We have used three real-world video anomaly datasets for experiments, namely UCF-Crime~\cite{sultani_chen_shah_2018}, CCTV-Fights~\cite{cctv_fights}, and UBI-Fights~\cite{ubi}.

\textbf{UCF-Crime~\cite{sultani_chen_shah_2018}:} It is a video anomaly dataset containing 13 real-world anomalies recorded using CCTV cameras. It contains 1900 real-world videos of normal and criminal activities such as robbery, vandalism, burglary, shooting, abuse, etc.


\textbf{CCTV-Fights~\cite{cctv_fights}}: This dataset offers 1000 fighting videos recorded in real-world scenarios. The total duration of these videos is 17.68 hours and collected using search keywords like kicking, punching, physical violence, mugging, etc.

\textbf{UBI-Fights~\cite{ubi}}: It holds 1000 real-world videos, where 784 are normal and 216 are real-life fighting scenarios. It contains videos recorded in indoor and outdoor environment with no administrative control or supervision, high occlusion, and varying illumination conditions.

\subsection{Performance Evaluation Metrics}
All test video frames from the above-aforementioned datasets are marked as either normal or abnormal. Hence following the previous works~\cite{mist,nguyen,ordinal,fastano,cctv_fights,sultani_chen_shah_2018,szyma_wacv22,xd_vio,jiaxingzhongnannanliweijiekong2019,Zhu2020Motion-awareDetection} on anomaly detection, we compute frame-level receiver operating characteristics (ROC) curve and area under the curve (AUC) as evaluation metrics.

\subsection{Comparisons with State-of-the-art}
We compare our method with recent state-of-the-art video anomaly detection methods~\cite{degardin2021iterative,continual,mist,mahmudul,Hinami2017JointKnowledge,ionescu,Kopuklu_2021_WACV, Leroux_2022_WACV,nguyen,ordinal,fastano,ramachandra,ravanbaksh_2018,sultani_chen_shah_2018,Zhu2020Motion-awareDetection} on three aforementioned datasets. Tab.~\ref{tab:sota_comparison} shows the performance of all methods. It can be observed that the proposed unsupervised method outperforms other weakly-supervised methods~\cite{mist,sultani_chen_shah_2018,Zhu2020Motion-awareDetection} by a substantial margin across all three datasets. Zhu~\cite{Zhu2020Motion-awareDetection}, Pang~\etal~\cite{ordinal}, and Leroux~\etal~\cite{Leroux_2022_WACV} have achieved decent performance on all datasets by introducing attention-based deep features, ordinal regression, and multi-branch deep autoencoders, respectively. However, incorporating multiple deep networks and adding attention-based features into the network are insufficient to detect multiple anomalous events. It can be observed that the multi-branch framework introduced by Leroux~\etal~\cite{Leroux_2022_WACV} performs well on CCTV-Fights~\cite{cctv_fights} and UBI-Fights~\cite{ubi} as these datasets focus on fighting events only. However, it performs moderately on UCF-Crime~\cite{sultani_chen_shah_2018} as the dataset addresses multiple anomalous activities. Doshi~\etal~\cite{continual} have employed continual learning in which the model learns new patterns as the input data arrives without forgetting the learnt information. However, such type of learning requires constant flow of incoming data. Moreover, such continual learning approach can efficiently utilize the temporal information of single fixed location~\cite{continual,ramachandra}. However, the chosen VAD datasets~\cite{ubi,cctv_fights,sultani_chen_shah_2018} for the experiments are multi-scene and provide complex temporal richness. To tackle this problem, Doshi~\etal~\cite{continual} have constructed NOLA video anomaly dataset using fixed location camera. However, to the best of our knowledge, this dataset is yet to be published. Perez~\etal~\cite{cctv_fights} have introduced CCTV-Fights dataset and computed the performance of C3D~\cite{d.tranl.bourdevr.fergusl.torresanim.paluri2017}, I3D~\cite{j.carreiraandrewzisserman2017}, and other popular backbone architectures. However, the popular 3D-CNN-based backbone architecture such as C3D~\cite{d.tranl.bourdevr.fergusl.torresanim.paluri2017} and I3D~\cite{j.carreiraandrewzisserman2017} have already been incorporated in the proposed framework as well as with other methods~\cite{mist,sultani_chen_shah_2018}. Hence we have not explicitly included the method used in~\cite{cctv_fights} for comparisons. However, we have studied the effectiveness of the backbone architectures in the proposed framework. Tab.~\ref{tab:backbone} shows AUC (in \%) for four popular architectures, namely Pseudo-ResNet 3D~\cite{p3d}, Temporal Segments Network~\cite{liminwangyuanjunxiong2016}, C3D~\cite{d.tranl.bourdevr.fergusl.torresanim.paluri2017}, and Inception V3~\cite{j.carreiraandrewzisserman2017}.

\begin{table}[tbh]
\centering
\caption{Frame-level AUC scores (in \%) of the state-of-the-art methods on three video anomaly datasets, D1: CCTV-Fights~\cite{cctv_fights}, D2: UBI-Fights~\cite{ubi}, and D3: UCF-Crime~\cite{sultani_chen_shah_2018}. The top two results are shown in red and blue.\\}
\label{tab:sota_comparison}
\resizebox{0.45\textwidth}{!}{%
\begin{tabular}{llllll}
\hline
\textbf{Year} & \textbf{Method}            & \textbf{D1} & \textbf{D2} & \textbf{D3}&Superv. \\ \hline
2016 & Hasan~\etal~\cite{mahmudul}       & 52.43       & 64.87      & 50.6 & Semi.    \\
2017 & Hinami~\etal~\cite{Hinami2017JointKnowledge}      & 56.70       & 67.12      & 57.10  & Semi.   \\
2018 & Ravanbaksh~\etal~\cite{ravanbaksh_2018}  & 60.37       & 69.45      & 61.61  & Unsuper.  \\
2018 & Sultani~\etal~\cite{sultani_chen_shah_2018}     & 72.55       & 78.70      & 75.41 & Weak.    \\
2019 & Ionescu~\etal~\cite{ionescu}     & 73.86       & 78.49      & 76.20  & Unsuper.   \\
2019 & Nguyen~\etal~\cite{nguyen}      & 76.43       & 77.18      & 75.65  & Semi.    \\
2019 & Zhu~\etal~\cite{Zhu2020Motion-awareDetection}         & 75.20       & 81.02      & 79.0 & Weak.      \\
2020 & Degardin~\etal~\cite{ubi}    & 77.14       & 84.60      & 76.90  & Weak.   \\
2020 & Ramachandra~\etal~\cite{ramachandra} & 73.81       & 82.45      & 75.46   & Semi. \\
2020 & Pang~\etal~\cite{ordinal}        & 76.78       & 84.65      & 78.50   & Unsuper.  \\
2021 & Feng~\etal~\cite{mist}        & \textcolor{red}{\textbf{81.43}}       & \textcolor{blue}{\textbf{85.19}}      & \textcolor{blue}{\textbf{82.30}}   & Weak.  \\
2021 & Kopuklu~\etal~\cite{Kopuklu_2021_WACV}     & 74.90       & 79.63      & 75.12   & Weak. \\
2022 & Doshi~\etal~\cite{continual}       & 75.86       & 80.71      & 79.46   & Semi. \\
2022 & Park~\etal~\cite{fastano}        & 73.28       & 77.23      & 75.40   & Unsuper.  \\
2022 & Leroux~\etal~\cite{Leroux_2022_WACV}      & 76.20       & 78.06      & 76.78  & Unsuper.   \\ \hline
     & \textbf{Ours} (Farneback Flow)  & 79.31       & 84.12      & 81.40   & Unsuper.  \\
     & \textbf{Ours} (SelFlow~\cite{selflow})    & \textcolor{blue}{\textbf{81.01}}       & \textcolor{red}{\textbf{86.31}}      & \textcolor{red}{\textbf{84.50}}  & Unsuper.   \\ \hline
\end{tabular}%
}
\end{table}

Following the limitations imposed by Zhong~\etal~\cite{jiaxingzhongnannanliweijiekong2019}, Pang~\etal~\cite{ordinal} have formulated anomaly detection as unsupervised ordinal regression and performed image-level anomaly detection. However, focusing only on spatial-features and ignoring temporal aspect of the video is not advisable in the context of anomalous event. Our framework utilizes both spatial and temporal information and hence outperforms the method proposed by Pang~\etal~\cite{ordinal} by a notable margin.

\begin{table}[tbh]
\centering
\caption{Performance of the proposed method in terms of AUC (\%) with different backbone architectures used for implementation of $\Omega$ and $\psi$ regressors.\\}
\label{tab:backbone}
\resizebox{0.37\textwidth}{!}{%
\begin{tabular}{llll}
\hline
\textbf{Backbone}  & \textbf{CCTV-Fight}~\cite{cctv_fights} & \textbf{UBI-Fight}~\cite{ubi} & \textbf{UCF-Crime}~\cite{sultani_chen_shah_2018} \\ \hline
P3D~\cite{p3d} & \textcolor{blue}{\textbf{78.42}}      & \textcolor{blue}{\textbf{84.20}}     & \textcolor{red}{\textbf{84.78}}     \\
TSN~\cite{liminwangyuanjunxiong2016}       & 77.10      & 83.08     & 81.22     \\
C3D~\cite{d.tranl.bourdevr.fergusl.torresanim.paluri2017}       & 76.56      & 81.91     & 79.96     \\
I3D~\cite{j.carreiraandrewzisserman2017}       & \textcolor{red}{\textbf{81.01}}      & \textcolor{red}{\textbf{86.31}}     & \textcolor{blue}{\textbf{84.50}}     \\ \hline
\end{tabular}%
}
\end{table}

Based on the performance results discussed so far, it is important to note the proposed framework i) addresses the feature selection problem faced by~\cite{mist,fastano,Zhu2020Motion-awareDetection} using low-level motion features and spatio-temporal features, ii) employs an iterative training rather than depending on the weak labels~\cite{sultani_chen_shah_2018,jiaxingzhongnannanliweijiekong2019,Zhu2020Motion-awareDetection}. Thus, our method has achieved a reasonable gain in terms of AUC (\%) score.

\subsection{Qualitative Analysis}
\begin{figure}[tbh]
    \centering
    \includegraphics[scale=0.55]{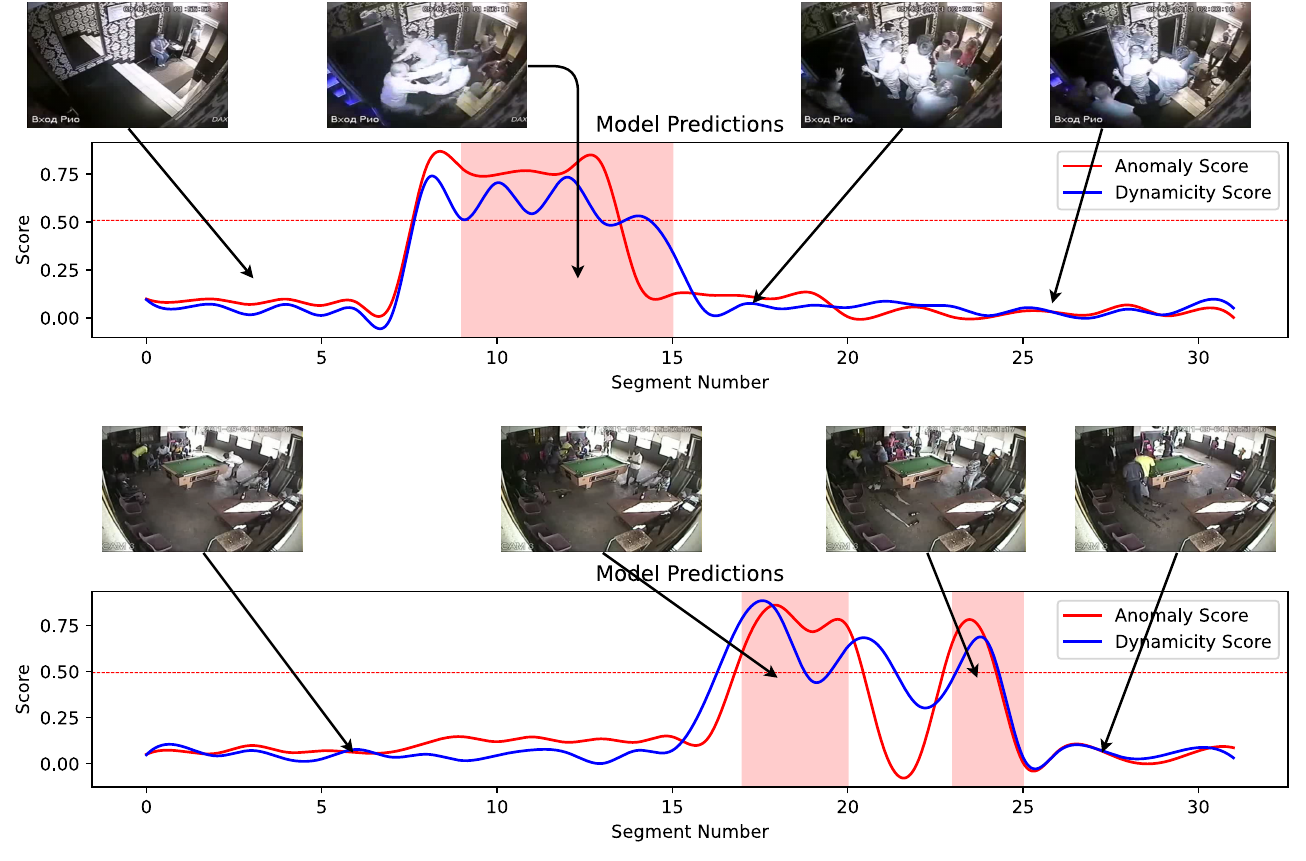}
    \caption{\textbf{Results Visualization}: Qualitative results on test videos taken from the CCTV-Fight~\cite{cctv_fights} dataset. Each image represents a frame in a temporal segment. The shaded portions are ground truths and the horizontal line represents the threshold.}
    \label{fig:visualization_cctv}
\end{figure}

\begin{figure}[tbh]
    \centering
    \includegraphics[scale=0.55]{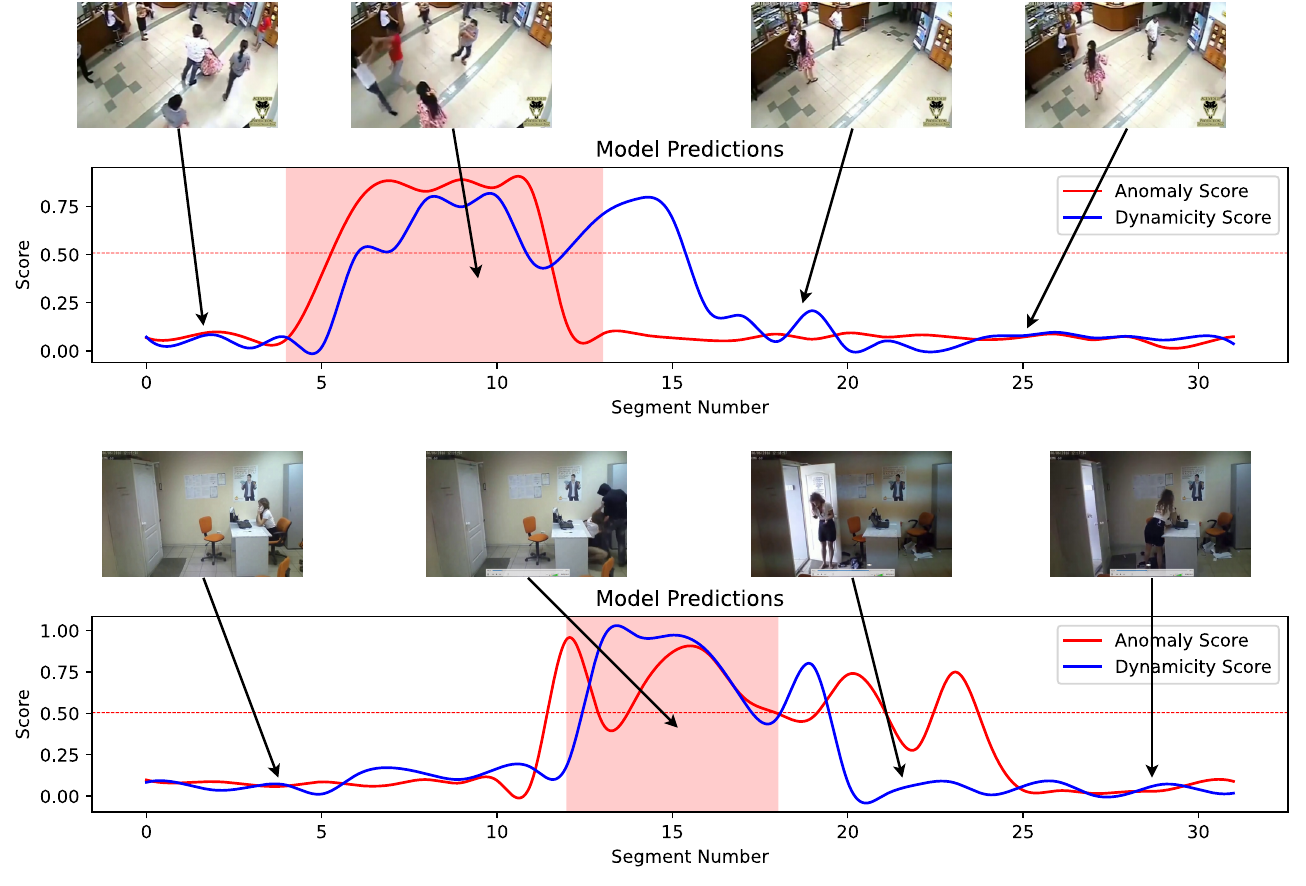}
    \caption{\textbf{Results Visualization}: Qualitative results on test videos taken from the UBI-Fight~\cite{ubi} dataset.}
    \label{fig:visualization_ubi}
\end{figure}

\begin{figure}[tbh]
    \centering
    \includegraphics[scale=0.55]{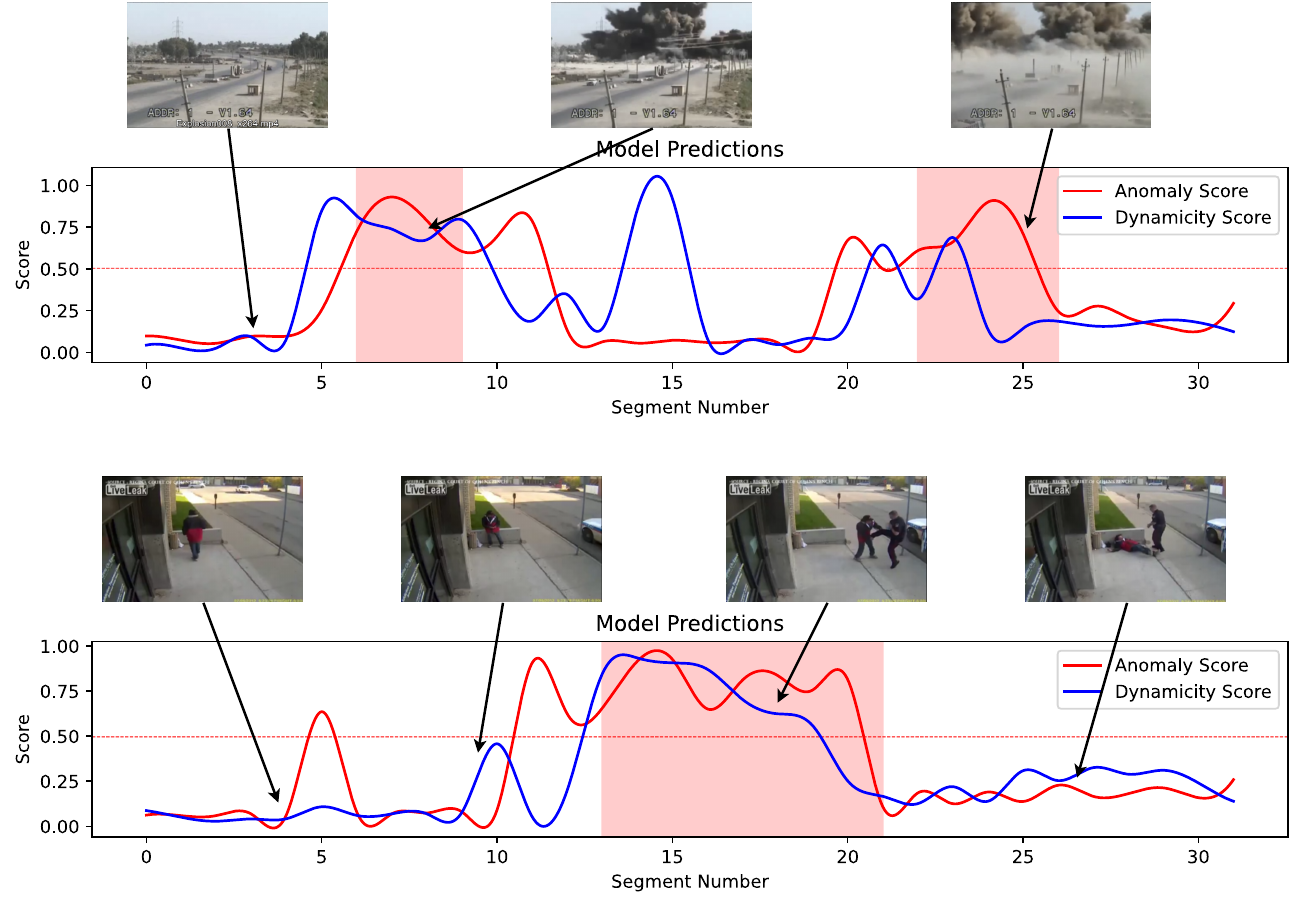}
    \caption{\textbf{Results Visualization}: Qualitative results on the test videos from the UCF-Crime~\cite{sultani_chen_shah_2018} dataset.}
    \label{fig:visualization_ucf}
\end{figure}

We present a few qualitative results obtained using the proposed method on a few test videos taken from  the CCTV-Fight~\cite{cctv_fights}, UBI-Fights~\cite{ubi}, and UCF-Crime~\cite{sultani_chen_shah_2018} datasets. Such results are presented in Figs.~\ref{fig:visualization_cctv},~\ref{fig:visualization_ubi}, and~\ref{fig:visualization_ucf}, respectively. Note that, the trained regressors $\Omega$ and $\psi$ generate corresponding segment-level anomaly and dynamicity scores. Hence we have interpolated these scores using Cubic Interpolation to achieve smooth curves. It can be seen that the method successfully detects anomalous segments
and generates higher anomaly and dynamicity scores as per the ground truths.

From Figs.~\ref{fig:visualization_cctv} and \ref{fig:visualization_ucf}, it can be seen that both regressors accurately detect anomalous patterns and abrupt change in the scene a few frames earlier. This indicates a quick response to sensitive contents. Moreover, the proposed framework is able to detect multiple occurrences of anomalous events in a video. From Figs.~\ref{fig:visualization_cctv} and \ref{fig:visualization_ubi}, it can be seen that in the absence of any anomalous activity, both regressors generate very low scores yielding lower false alarms toward the later part of the videos.

In Fig.~\ref{fig:visualization_ucf}, the first illustration depicting an explosion event from the UCF-Crime~\cite{sultani_chen_shah_2018} dataset is very interesting. The explosion usually fills the whole field of view of the camera with a thick smoke that moves slowly or rapidly depending on the intensity of the explosion event. In this example, after successfully detecting the first explosion, due to faster moving smoke, the $\psi$ regressor has generated a high dynamicity score. However, detecting smoke is not necessarily an anomalous event. Hence $\Omega$ has predicted a very low anomaly score for the same segment avoiding false positives. However, during the second mild-level explosion, both regressors agree to generate a relatively higher score. More qualitative
results on anomaly detection have been provided in the supplementary material.

\subsection{Number of Passes and Training Iterations}
\begin{figure}[tbh]
    \centering
    \includegraphics[scale = 0.39]{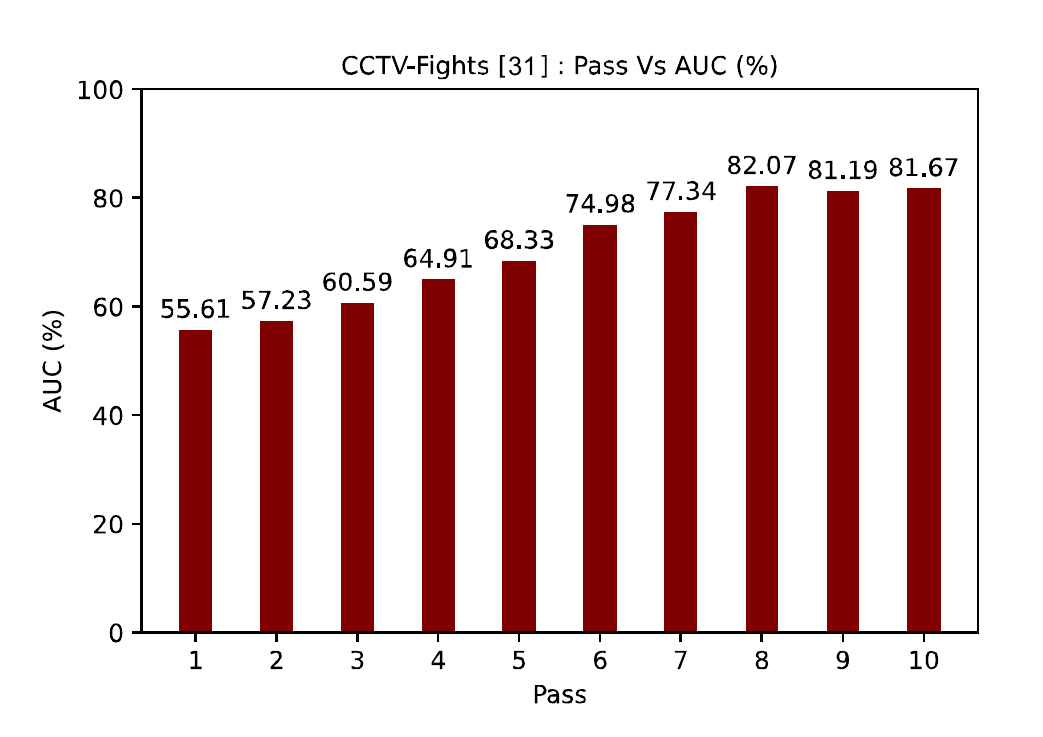}
    \caption{\textbf{Pass Vs. AUC}: The AUC (in \%) performance of the proposed method for the CCTV-Fights~\cite{cctv_fights} dataset videos against each pass, where x-axis is the number of passes and y-axis presents AUC.}
    \label{fig:cctv_bars}
\end{figure}

\begin{figure}[tbh]
    \centering
    \includegraphics[scale = 0.39]{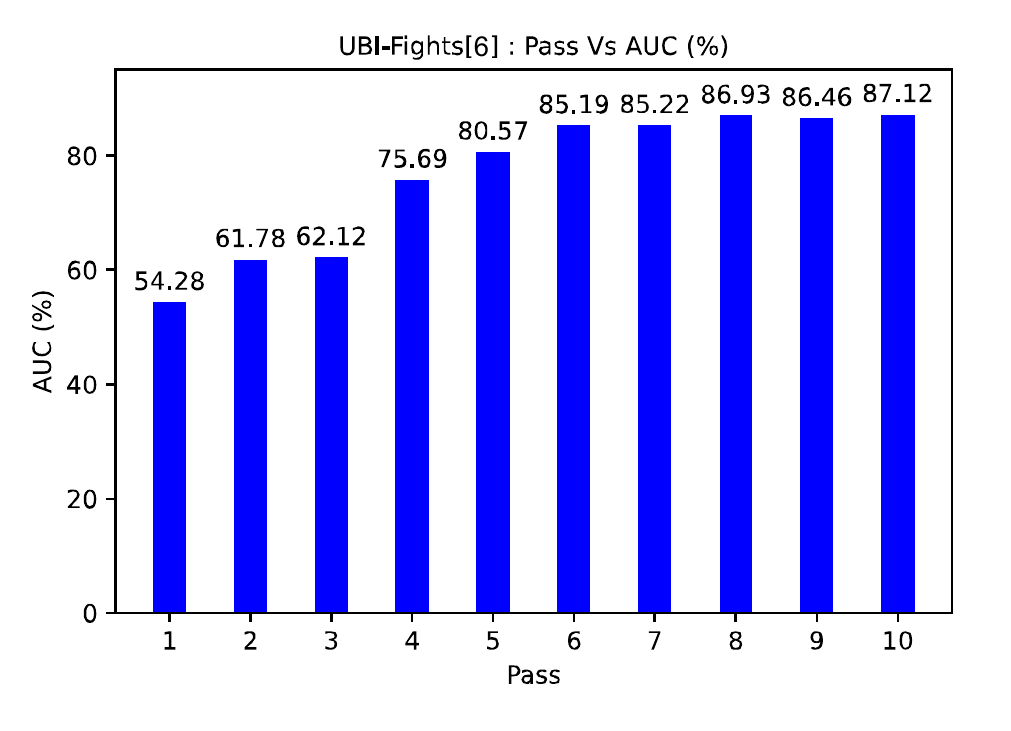}
    \caption{\textbf{Pass Vs. AUC}: The AUC (in \%) performance of the proposed method for the UBI-Fights~\cite{ubi} dataset videos against each pass.}
    \label{fig:ubi_bars}
\end{figure}

\begin{figure}[tbh]
    \centering
    \includegraphics[scale = 0.39]{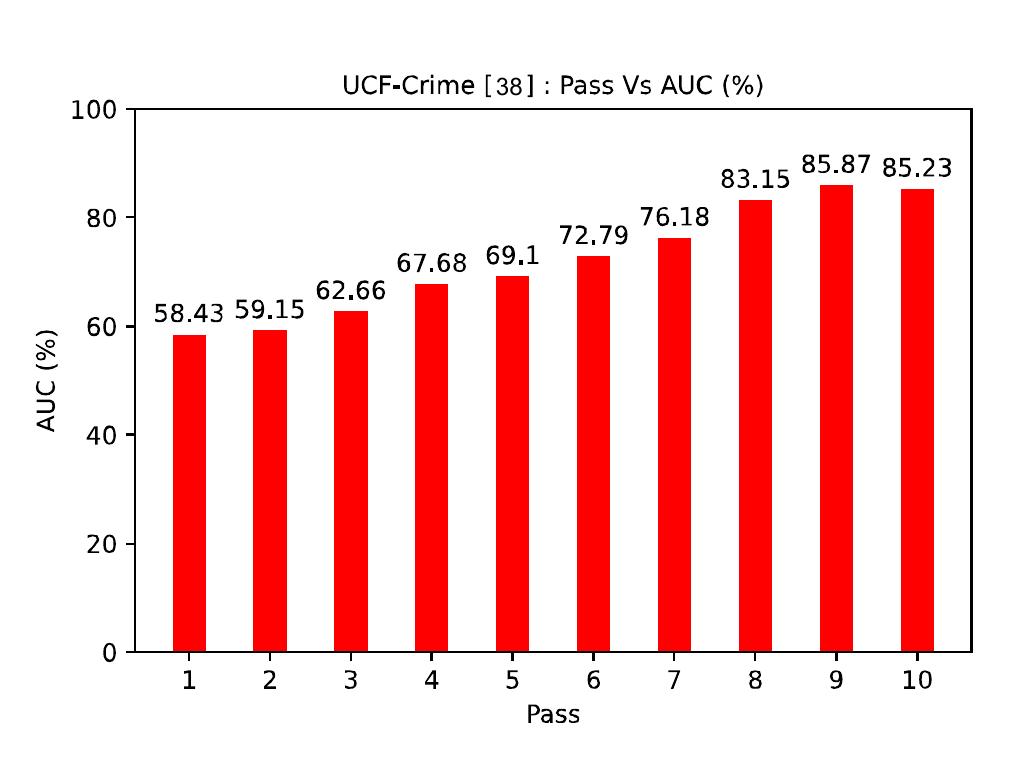}
    \caption{\textbf{Pass Vs. AUC}: The AUC (in \%) performance of the proposed method for the UCF-Crime~\cite{sultani_chen_shah_2018} dataset videos against  each pass.}
    \label{fig:ucf_bars}
\end{figure}

To understand the iterative training mechanism, we present the AUC (in \%) results of the proposed framework at each pass during the training on CCTV-Fight~\cite{cctv_fights}, UBI-Fights~\cite{ubi}, and UCF-Crime~\cite{sultani_chen_shah_2018} in Figs.~\ref{fig:cctv_bars}, \ref{fig:ubi_bars}, and \ref{fig:ucf_bars}. For CCTV-Fights~\cite{cctv_fights}, our method achieves stable performance at the 9th and 10th pass. However, for UBI-Fights~\cite{ubi} and UCF-Crime~\cite{sultani_chen_shah_2018}, the framework's performance has improved  significantly in the first few passes across all datasets. It achieves a stable performance after the 7th or 8th pass. Note that, during each pass, the sub-optimized version of $\Omega$ and $\psi$ is retrained with the refined pseudo labels. Hence it is necessary to restricts this training to avoid over-fitting. We have observed that the number of training iterations can be decided by the input batch size and the total number of samples in the training set. For example, CCTV-Fight~\cite{cctv_fights}, UBI-Fights~\cite{ubi}, and UCF-Crime~\cite{sultani_chen_shah_2018} datasets contain a few thousands of video samples in the training set. Hence 30 training iterations/pass with a batch size 32 is sufficient to train the model. However, we have experimentally found that the number of iterations/pass do not matter much as long as large number of iterations are done within a pass. This  ensures that the models are retrained iteratively. Hence we can achieve same performance with fewer number of training iterations and a large number of passes and vise-versa. Since all datasets offer a few thousands of samples in the training set, we have found that 10 passes and 30 training iterations are sufficient to train both the regressors. All experiments in this paper have thus been conducted under this uniform setting. 

\subsection{Ablation Study}
The proposed method has three main modules: i) pseudo label assignment, ii) backbone architecture, and iii) dynamicity score to efficiently detect anomalies. In the first stage, we have employed OneClassSVM and iForest~\cite{isolation} to obtained pseudo anomaly score for each temporal segment. We have replaced these two unsupervised anomaly detection algorithms with Robust Covariance~\cite{robust} and Local Outlier Factor (LOF)~\cite{lof}. However, a significant degradation in the AUC performance has been observed (3\% - 5\%). In the second stage, we have employed two-stream I3D~\cite{j.carreiraandrewzisserman2017} followed by a 3-layer FCN to generate the scores. To check the efficiency of this backbone, we have re-conducted the experiments with same setting. The overall AUC performance with respect to the backbone is represented in Tab.~\ref{tab:backbone}. We have also represented the effect of considering low-level motion features to decide the abnormality of the scene. From Tab.~\ref{tab:wo_dynamicity} and qualitative results, it can be safely concluded that, inclusion of motion features helps to achieve good detection performance as well as lower False Alarms Rate (FAR). We have explored various unsupervised algorithms to generate pseudo anomaly scores. Tab.~\ref{tab:pseudo_ablation} presents the AUC performance of these experiments. It reveals that when OCSVM is combined with iForest, we get best performance.

\begin{table}[tbh]
\centering
\caption{AUC (in \%) and False Alarms Rate (FAR) of the proposed method with and without the dynamicity score. An improved AUC and corresponding FAR are shown in red and blue colors, respectively. \\}
\label{tab:wo_dynamicity}
\resizebox{0.40\textwidth}{!}{%
\begin{tabular}{llll}
\hline
\textbf{Dynamicity} & \textbf{CCTV-Fight}~\cite{cctv_fights}  & \textbf{UBI-Fights}~\cite{ubi}  & \textbf{UCF-Crime}~\cite{sultani_chen_shah_2018}   \\ \hline
No         & 75.21 (5.8) & 81.64 (4.7) & 79.76 (1.8) \\
Yes        & \textcolor{red}{\textbf{81.01}} \textcolor{blue}{\textbf{(1.7)}} & \textcolor{red}{\textbf{86.31}} \textcolor{blue}{\textbf{(1.4)}} & \textcolor{red}{\textbf{84.50}} \textcolor{blue}{\textbf{(0.5)}} \\ \hline
\end{tabular}%
}
\end{table}

\begin{table}[tbh]
\centering
\caption{Performance of the proposed method in terms of AUC (\%) with different unsupervised algorithms combined with iForest~\cite{isolation} to generate pseudo-anomaly scores.\\}
\label{tab:pseudo_ablation}
\resizebox{0.37\textwidth}{!}{%
\begin{tabular}{llll}
\hline
\textbf{Algorithm} & \textbf{CCTV-Fights}~\cite{cctv_fights} & \textbf{UBI-Fights}~\cite{ubi} & \textbf{UCF-Crime}~\cite{sultani_chen_shah_2018} \\ \hline
MCD       & 77.24       & 84.07      & 81.11     \\
PCA       & \textbf{\textcolor{blue}{79.94}}       & \textbf{\textcolor{blue}{85.13}}      & \textbf{\textcolor{blue}{83.58}}     \\
LOF       & 77.60       & 84.86      & 82.02     \\
OCSVM     & \textbf{\textcolor{red}{81.01}}       & \textbf{\textcolor{red}{86.31}}      & \textbf{\textcolor{red}{84.50}}     \\ \hline
\end{tabular}%
}
\end{table}

\section{Conclusion and Future Work}
\label{section:conclusion}

It has been discussed in this paper that large-scale video anomaly detection using iterative learning is a viable approach to avoid annotation dependency. We have shown that by employing  iterative training, the model can  learn discriminating features. Moreover, we have shown that by employing pseudo-label generation, one can avoid any type of supervision and still achieve very good performance. Two key insights are: i) low-level features are equally important for anomaly detection, and ii) iterative training helps to reduce FAR and it is possible to detect anomalous event a few frames earlier. We can explore more advance technique to utilize both low-level and deep features in future. However, it is not wise to assume that any AI assisted visual surveillance framework can be a complete replacement of manual surveillance. Essentially, the amount of training data and quality of the underlying model play important role in decision making. 

\section*{Acknowledgement}
This work was supported in part by the Korea Institute of Science and Technology (KIST) Institutional Program under Project 2E31082 and in part by the National Research Foundation (NRF) Project (Grant No. 2018M3E3A1057288) executed at IIT Bhubaneswar with project code CP220.
{\small
\bibliographystyle{ieee_fullname}
\bibliography{egbib}
}

\end{document}